\documentclass{article}

    \usepackage[preprint]{neurips_2020}

\usepackage{natbib}
\usepackage[utf8]{inputenc} 
\usepackage[T1]{fontenc}    
\usepackage[colorlinks=true,linkcolor=blue,citecolor=blue]{hyperref}       
\usepackage{url}            
\usepackage{booktabs}       
\usepackage{amsfonts}       
\usepackage{nicefrac}       
\usepackage{microtype}      
\usepackage{amsmath}
\usepackage{graphicx}

\newcommand{\norm}[1]{\left\lVert#1\right\rVert}

\title{Robust Face Verification via Disentangled Representations}

\author{%
  Marius Arvinte \\
  ECE Department \\
  UT Austin\\
  \texttt{arvinte@utexas.edu} \\
   \And
   Ahmed H. Tewfik \\
   ECE Department \\
   UT Austin\\
   \texttt{tewfik@austin.utexas.edu} \\
   \And
   Sriram Vishwanath \\
   ECE Department \\
   UT Austin\\
   \texttt{sriram@ece.utexas.edu} \\
}

\begin{document}

\maketitle

\begin{abstract}
We introduce a robust algorithm for face verification, i.e., deciding whether two images are of the same person or not. Our approach is a novel take on the idea of using deep generative networks for adversarial robustness. We use the generative model during training as an online augmentation method instead of a test-time purifier that removes adversarial noise. Our architecture uses a contrastive loss term and a disentangled generative model to sample negative pairs. Instead of randomly pairing two real images, we pair an image with its class-modified counterpart while keeping its content (pose, head tilt, hair, etc.) intact. This enables us to efficiently sample hard negative pairs for the contrastive loss. We experimentally show that, when coupled with adversarial training, the proposed scheme converges with a weak inner solver and has a higher clean and robust accuracy than state-of-the-art-methods when evaluated against white-box physical attacks. Source code is available at \url{https://github.com/mariusarvinte/robust-face-verification}.
\end{abstract}

\section{Introduction}
Deep neural networks have become the \textit{de facto} model of choice for complex computer vision tasks such as large-scale image classification \citep{russakovsky2015imagenet}, face recognition \citep{Parkhi15, liu2017sphereface}, high quality image synthesis \citep{karras2019analyzing}, and denoising \citep{zhang2017beyond}. In particular, face recognition, re-identification, and verification have seen marked improvements when paired with deep learning \citep{masi2018deep}. At the same time, adversarial examples \citep{goodfellow2014explaining} have demonstrated to be a failure mode of deep neural networks not only in the digital domain but in the physical world as well \citep{kurakin2016adversarial, sharif2016accessorize, pmlr-v80-athalye18b}.

Adversarial training \citep{goodfellow2014explaining, madry2018towards} is one of the more successful empirical defense methods that has inspired numerous extensions \citep{zhang2019towards, xie2019feature}. Our work is carried out in this direction and introduces an architecture for adversarially robust face verification against \textit{physical attacks}. Our proposed method trains an embedding model using a similarity loss, a generative model to augment pairs of samples and a \textit{weak} adversarial training procedure --- where the location (but not the pattern) of the perturbation is chosen randomly.

As a motivating example, consider the following scenario, where we train a robust model that recognizes whether two images are of the same person or not. During training, we sample a negative pair of images and add an adversarial eye patch to one of them. If the two samples contain persons facing different directions, then pose also acts as a discriminator and prevents the model from learning robust features. If we would have pairs of samples that are \textit{aligned} in terms of content (pose, background, smile, etc.), then the network would focus on learning features that are meaningful when obscured by the patch. In our proposed model we use a generative model with disentangled latent factors as a form of online data augmentation to generate such pairs during training by \textit{transferring class information between samples}.

Our main contribution is, to the best of our knowledge, the first architecture that uses a generative model for data augmentation to reach convergence when trained with a \textit{weak} inner adversary. Our approach randomly selects a perturbation mask at each step. We show that this leads to higher clean \textit{and} robust accuracies over state-of-the-art methods, while being more computationally efficient. State-of-the-art approaches implement a non-augmented model trained using a near-optimal inner adversary. Furthermore, we find that non-augmented adversarial training with random patch locations -- as approximators to physical attacks -- does not converge to a robust solution and instead exhibits a critical overfitting phenomenon \citep{rice2020overfitting}. 


\section{Proposed Method}
Let $G(z_\mathrm{cl}, z_\mathrm{co})$ be a deep generative model with a partitioned latent space, such that $z_\mathrm{cl}$ is responsible for modulating the class identity of the generated image $x$, whereas $z_\mathrm{co}$ carries all information about the content. Let $E_{cl}(x)$ and $E_{co}(x)$ be two deep neural networks (encoders) that extract the class and content information from $x$, respectively. Together, the composed model $G(E_{cl}(x), E_{co}(x))$ can be viewed as an autoencoder with a disentangled latent space. A formal notion of disentanglement is still an open problem, but we use the recent definitions introduced in \cite{Shu2020Weakly}: we consider a generator disentangled if modifying one of the two components does not change the other when re-extracted from the new sample by an oracle function. While we do not have a closed-form expression for the oracle functions, we think of them as equivalent to human-level perception.

\begin{figure}
    \centering
    \includegraphics[width=0.9\textwidth]{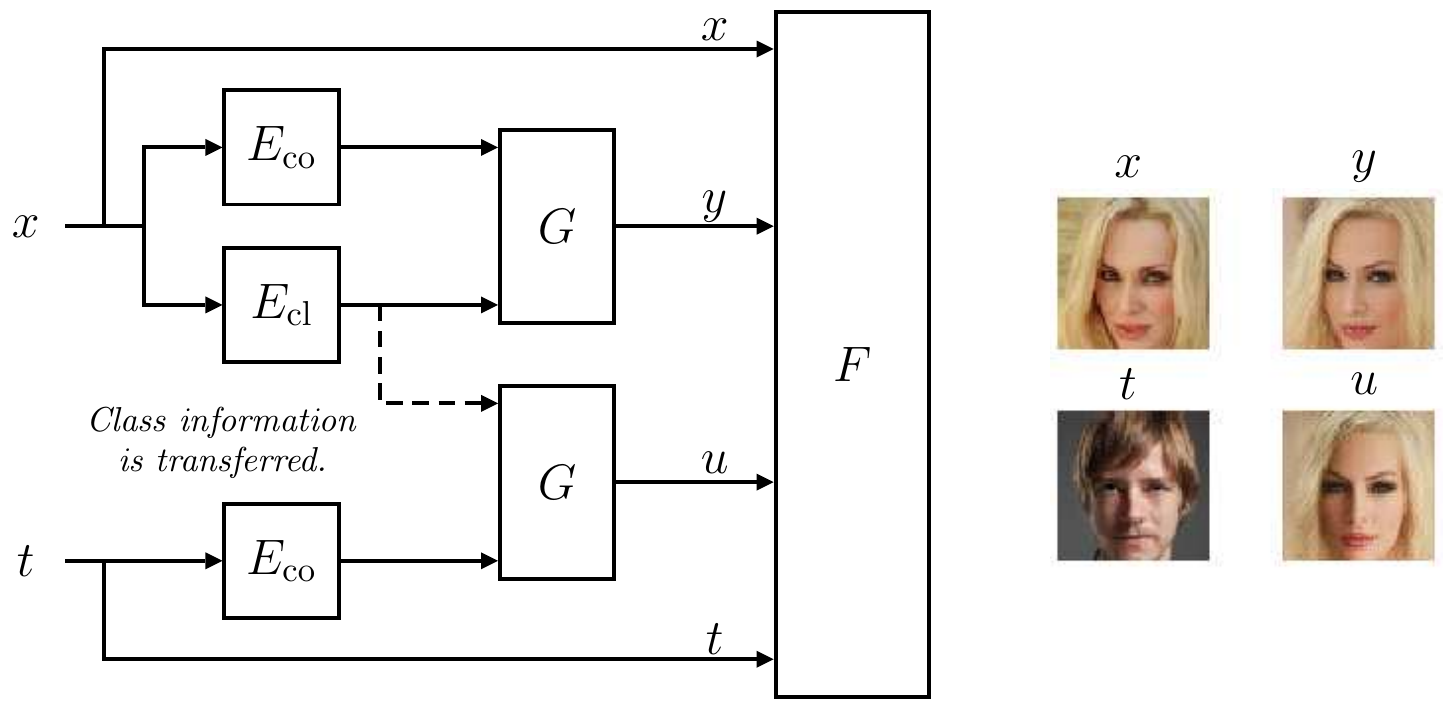}
    \caption{\textbf{Left}: Architecture of the proposed approach. Blocks containing the same sub-models are identical (weights are shared at all times). \textbf{Right}: Example of $x, y, t, u$ quadruplets.}
    \label{fig:block_diagram}
\end{figure}

Let $x$ and $t$ be two samples with potentially different ground truth class identities $c_x$ and $c_t$. Our goal is to learn a binary decision function $D$ that takes as input the pair $(x, t)$ and returns one if $c_x \neq c_t$ and zero otherwise. A zero output signifies $D$ perceiving that $x$ and $t$ have the same class identity. We use the feature embedding from a deep neural network $F$ and rewrite the detector that uses $x$ as \textit{target} and $t$ as \textit{candidate} image as

\begin{equation}
    D(t; x) =
    \begin{cases}
          1, & \norm{F(t) - F(x)}_2 \geq \delta \\
          0, & \text{otherwise} \\
    \end{cases}
\end{equation}

\noindent where $\delta$ is a detection threshold and $x$ has class identity $c_x$, assumed to be unaltered by any adversary. In practice, for a face verification system, $x$ can be a single image taken of the target person in a physically secure environment. To efficiently train the embedding function $F$, we propose a \textit{two-pair contrastive loss} term \citep{hadsell2006dimensionality}, in the form of

\begin{equation}
L(x, y, t, u) = \norm{F(x) - F(y)}_2^2 + \max\{ m - \norm{F(t)-F(u)}_2^2, 0 \}
\end{equation}

\noindent where $m$ is the margin between the positive and negative pairs. A block diagram is shown in Figure \ref{fig:block_diagram}. $x, y, t, u$ are four distinct samples such that $x, y, u$ share the same class identity $c_x$ and $t$ has a different class identity $c_t \ne c_x$.

The four samples are obtained accordingly:
\begin{itemize}
    \item $x$ and $t$ are sampled independently from the underlying distributions governing $c_x$ and $c_t$, respectively.
    \item $y = G \left( E_\mathrm{cl}(x), E_\mathrm{co}(x) \right)$. That is, $y$ is the pass-through reconstruction of $x$ through the autoencoder.
    \item $u = G \left( E_\mathrm{cl}(x), E_\mathrm{co}(t) \right)$. That is, $u$ is obtained by generating a sample with the same class identity as $x$ but with the content information of $t$.
\end{itemize}

In practice, we only have access to a finite number of labeled samples for each class identity, each with different content information, e.g., different pictures of the same person in different poses. To sample $x$ and $t$, we first randomly sample a class $c_x$, then uniformly sample in- and out-of-distribution images. In the non-robust version, the weights $\theta$ of the network are shared and receive gradient updates from both pairs simultaneously towards the objective function

\begin{equation}
\min_{\theta} \mathbb{E}_{x, t} \left[ L(x, y, t, u; \theta) \right]
\end{equation}

\noindent where $\theta$ are the weights of $F$ optimized with stochastic gradient descent.

\subsection{Why Autoencode $x$?}

Notice that, if the autoencoder given by $G(E_\mathrm{cl}, E_\mathrm{co})$ has zero generalization error, then $y = x$ and $F(y) = F(x)$. This would lead to a collapse of our loss function, since it would only learn to maximize the loss between different samples, encouraging the learning of a chaotic mapping that does not actually take the person's identity into account. There are two key factors that help regularize against this type of collapse, outlined as follows.

\paragraph{The autoencoder is not ideal.} The reconstruction error is distributed asymmetrically between the class and content information. While there are slight variations in the class identity, the background is often reconstructed poorly, as shown for example in \cite{bau2019seeing}. We find that this also holds true for the generative models that we use in our experiments. This helps the embedding model by signaling that background information is not relevant for good discrimination.

\paragraph{Data augmentation.} We perform patch augmentation \citep{devries2017improved, lopes2019improving}, followed by patch adversarial training on the real images $x$ but not on the synthetic images $y$ and $u$. This encourages the embedding $F$ to learn robust features that signal facial similarity.

\subsection{Why Generate $u$?}

The main purpose of $G$ in the proposed scheme is to efficiently generate negative pairs closer to the decision boundary by transferring all content information between classes, allowing $F$ to learn discriminative features regarding the class identity. In particular, for face images, the learned features should be independent of the person's pose, expression, etc. This role is played even in the absence of any augmentation in the negative pair and is further magnified by adversarial training.

Together, the generated images $y$ and $u$ serve as virtual anchors for learning the embedding $F$. Furthermore, since they both have the same class identity, our proposed loss function is similar to a triplet loss, except that the anchor is different in the two pairs, but preserves class information. It could be possible to replace $y$ with a real sample with different content information, however, this causes the learning to quickly overfit, since $u$ is the only sample produced by the generator and has spurious artifacts that act as discriminators.

\subsection{Robust Optimization}
We adopt the framework of \cite{madry2018towards} and define the robust optimization objective as

\begin{equation}
    \min_{\theta} \mathbb{E}_{x, t, \mathcal{M}_x, \mathcal{M}_t} \left[ \max_{\delta_x, \delta_t} L \left( x+\mathcal{M}_x \cdot \delta_x, \mathrm{sg}(y), t+\mathcal{M}_t \cdot \delta_t,  \mathrm{sg}(u) ; \theta \right) \right]
\end{equation}

The variables $\mathcal{M}_x$ and $\mathcal{M}_t$ represent binary masks that restrict the perturbation to a subset of input pixels. Critical to our approach is the fact that we \textit{randomly} sample $\mathcal{M}_x$ and $\mathcal{M}_t$, instead of including them in the inner optimization objective. This allows us to reduce computational complexity but also significantly weakens the online adversary. The function $\mathrm{sg}$ represents the stop gradient operator, meaning that $y$ and $u$ are treated as constants for the inner optimization problem. Expanding the loss function allows us to decompose the optimization over both inputs as

\begin{equation}
\min_{\theta} \mathbb{E}_{x, t, \mathcal{M}_x, \mathcal{M}_t} \left[ \max_{\delta_x} \norm{F ( x+\mathcal{M}_x \cdot \delta_x) - \mathrm{sg}(F(y))}_2^2 - \min_{\delta_t} \norm{F(t+\mathcal{M}_t \cdot \delta_t) - \mathrm{sg}(F(u))}_2^2  \right]
\end{equation}

The first term can be viewed as a form of untargeted optimization: we optimize $\delta_x$ such that $x$ loses its original class identity (when viewed through $F$). Meanwhile, the second term is a form of targeted optimization: $\delta_t$ is optimized such that $t$ is viewed as belonging to a particular class identity. Note that we neglect $m$ in the above for brevity. From this formulation we derive another form of robust optimization that operates with real (non-augmented) image pairs and we call \textit{weak adversarial training}. Letting $u=y$ and sampling $y$ from the distribution of real images, we effectively remove the generator and both encoders from the optimization, recovering the optimization problem

\begin{equation}
   \min_\theta \mathbb{E}_{x, y, t, \mathcal{M}_x, \mathcal{M}_t} \left[ \max_{\delta_x} \norm{F(x+\mathcal{M}_x \cdot \delta_x) - F(y)}_2^2  - \min_{\delta_t} \norm{F(t+\mathcal{M}_t \cdot \delta_t) - F(y)}_2^2  \right]
   \label{eq:weak_at}
\end{equation}

\section{Physical Adversarial Attacks and Defenses}
We use mask-based attacks as an approximation for physically realizable attacks, similar to \cite{Wu2020Defending}. That is, we assume that an adversary can modify a contiguous region of the image without any restrictions on the norm of the perturbations. The two degrees of freedom an adversary has are related to the relative size of the perturbed region $\mathcal{M}$ and its location in the image.

We assume that the adversary has complete (white-box) knowledge of the detector $D$, including the weights and architecture of $F$, as well as the image of the target person $x_t$ and their corresponding features $F(x_t)$. We test our approach against impersonation attacks (an adversary starts from an image $x$ with class identity different than $c_t$ and attempts to bypass the detector $D$ so as to label them as $c_t$) and evasion attacks (the adversary attempts to evade identification). Given a white-box impersonating adversary and a fixed position for the allowed perturbations as $\mathcal{M}$, the optimization objective is straightforward to formalize as an unconstrained feature adversary \citep{sabour2015adversarial}

\begin{equation}
\min_{\delta} \norm{F(x+\mathcal{M} \cdot \delta) - F(x_t)}_2^2.
\label{eq:feature_adv}
\end{equation}

We evaluate our performance against the following threat models:

\begin{enumerate}
    \item Adversarial eyeglass frame attack with a fixed location for the mask $\mathcal{M}$ covering $2.6\%$ of the image.
    \item Square patch attack \citep{Wu2020Defending} covering $2.5\%$ of the image, in which the attacker performs a search to find the best patch location.
    \item Universal eye patch attack \citep{brown2017adversarial} covering $10\%$ of the image, where a single adversarial pattern with a fixed location covering the eyes is trained for multiple intruders.
\end{enumerate}

The adversarial eyeglasses and square patch are single-image attacks for which an instantaneous perturbation is found, thus are only an approximation for physical conditions. We do not consider the expectation-over-transformation framework \citep{pmlr-v80-athalye18b}, which imposes more restrictions on the attacker. Doing this makes our attacks stronger, on average, but occupying a smaller area of the image.

\subsection{Adversarial Training against Occlusion Attacks}

\begin{figure}
    \centering
    \includegraphics[width=1.\textwidth]{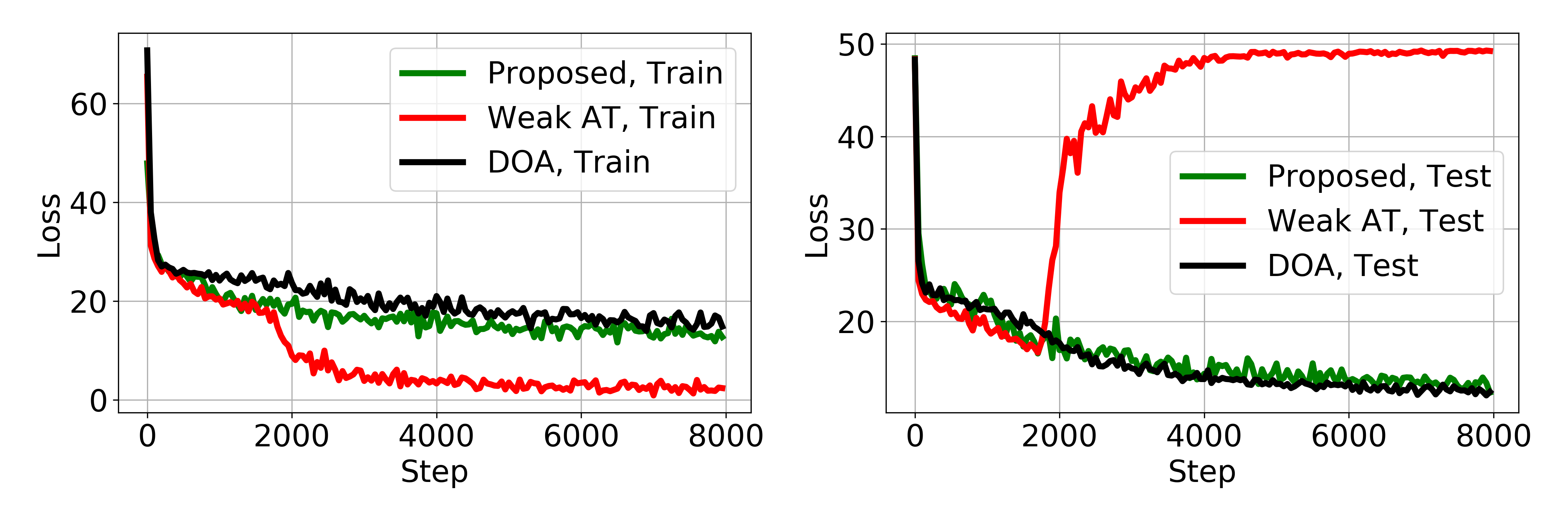}
    \caption{Train/test loss during training for the three proposed methods, with the same hyper-parameters. Weak adversarial training on real images shows overfitting, while our approach converges to a robust solution.}
    \label{fig:train_dynamics}
\end{figure}

\cite{Wu2020Defending} propose to extend the robust optimization framework of \cite{madry2018towards} to defend against occlusion attacks. This poses another important question: how to choose the patch location $\mathcal{M}$ during training? This is an issue, since the ideal solution would be to perform the attack for all patch locations and then pick the best result. However, even simply computing the initial loss function at all possible patch locations is computationally prohibitive, since it requires a large amount of forward passes each batch. To solve this issue, \cite{Wu2020Defending} propose an approximate search algorithm that integrates the magnitude of the gradient across different patch locations, picks the top $C$ candidates, and finally selects the location with the highest value of the loss function after in-painting.

We ask the following question: \textit{is it possible to train a robust classifier by randomly selecting the occlusion mask at each optimization step}? As baselines, we consider weak patch adversarial training directly on the triplet loss in (\ref{eq:weak_at}) with random mask selection and the gradient-based selection in \cite{Wu2020Defending}.

\section{Experimental Results}

We use the disentangled generative model introduced by \cite{Gabbay2020Demystifying}, trained on the CelebA dataset \citep{liu2015faceattributes}. We define the class code as the person identity and the content code as all other information included in the image and split the images in 9177 training persons and 1000 test persons. The cost of training $G$ is approximately $14$ hours for $200$ epochs. For the embedding network, we use the architecture in VGGFace \citep{Parkhi15} with the following modifications: we normalize the weights of the final embedding layer to unit $L_2$-norm and we remove the ReLU activation to help regularize the model during training and enable a more expressive feature space. For the two-pair contrastive loss, we set $m = 10$.

We train $F$ using a subset of images corresponding from 2000 samples from the data used to train the generative model and validate the performance on a set of 100 test persons. The test persons are never seen by $G, E_\mathrm{cl}, E_\mathrm{co}$ or $F$. We use the Adam optimizer \cite{kingma2014adam} with learning rate $0.001$ and PGD-based adversarial training \citep{madry2018towards} with a constraint of $\epsilon = 1$. We perform a hyper-parameter search for the best combination of step size and number of steps, and find that $10$ steps with step size $16/255$ produces the best results. For our approach, we measure validation loss exclusively on real pairs of images. For the feature adversary, we run an Adam optimizer on (\ref{eq:feature_adv}) for $1000$ iterations with learning rate $0.01$ and $5$ restarts. For the square patch attack we search for the best location by running $20$ iterations with a larger learning rate across all possible locations. 

Training takes approximately 11 and 25 hours for our method (excluding the cost of training $G$) and DOA, $C=10$, respectively, on a single Nvidia RTX 2080Ti GPU.

\subsection{Weak Adversarial Training}

\begin{figure}
    \centering
    \includegraphics[width=1.\textwidth]{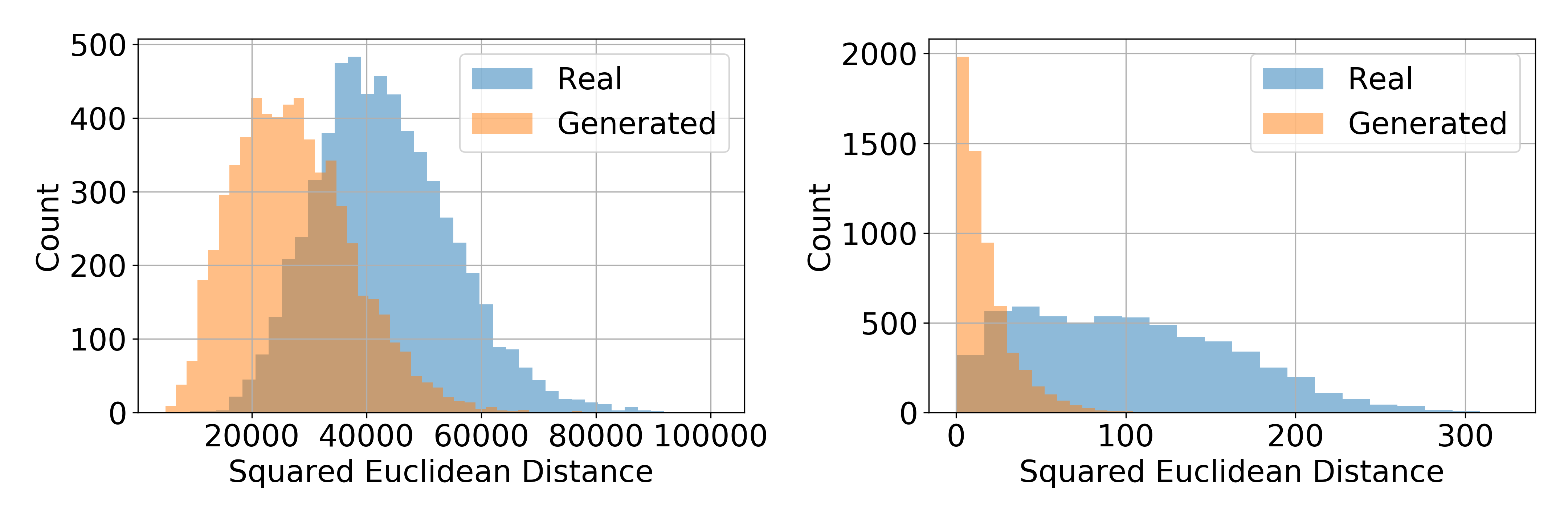}
    \caption{Feature distance between a pair of real-real and real-generated images with different class identities. \textbf{Left}: Measured with a pretrained VGGFace \citep{Parkhi15} network. \textbf{Right}: Measured with a robust network trained with DOA \citep{Wu2020Defending}.}
    \label{fig:measuring_similarity}
\end{figure}

Our first result is shown in Figure \ref{fig:train_dynamics} and indicates that the proposed approach convergences with random patch locations, whereas training on real image pairs does not. This validates our approach as an online data augmentation method that enables fast converge to a robust solution \citep{rice2020overfitting}. Furthermore, even when considering the Defense against Occlusion Attacks (DOA) approach as a baseline, which performs an approximate search for the best patch location, our approach shows comparable test performance while being less expensive.

The divergence of weak adversarial training is reminiscent of mode collapse in generative adversarial networks, where the generator overpowers a discriminator that is weakly trained (either through the number of steps or the architectural choice) and the loss function collapses, implying that randomly selecting a patch location significantly weakens the inner adversary. To illustrate the role of $G$ as a helper to the inner adversary Figure \ref{fig:measuring_similarity} plots the pairwise distance between persons with different class identities when the second image in the pair is either real or generated. To measure this distance we use two networks, independent of our approach: a pretrained VGGFace network \citep{Parkhi15} and a robust network trained with DOA \citep{Wu2020Defending}, since prior work \citep{ilyas2019adversarial} has shown that robust networks serve as good feature extractors.

Furthermore, even though the test losses are comparable, our approach achieves better test areas under the receiver operating characteristic curve (AU-ROC) and precision-recall (AU-PR) curve, thus is superior for face verification purposes. We use a weak form of semi-hard example mining to further regularize the convergence of all methods. For each negative pair, we sample a number of two persons with identity different than the target and pick the one that produces a lower feature distance when matched against the target. In the case of our approach, we first reconstruct both candidate intruders and measure similarity between $t$ and $u$, instead of comparing them with the real image $x$.

\subsection{Test Time Augmentations}
We augment $F$ at test time with face mirroring, which is routinely used to boost the performance of face recognition systems \citep{schroff2015facenet}. That is, we extract representations from both the candidate image and its symmetrically flipped version and then compute the average feature distance with respect to the ones extracted (or stored) of the target image. Additionally, when we have multiple samples of the target available, we perform target sample selection: we choose as target the image which minimizes the average feature space distance to all others. In practice, this means choosing a clear, non-obscured image as the target, as opposed to outliers. We apply these augmentations to both our baselines.

We test our approach on a set of samples corresponding to $27$ randomly selected test persons from the CelebA dataset, each with at least ten samples. The results are shown in Table \ref{table:results}, from which we see that our approach is superior to both weak adversarial training (with early stopping at the best validation AU-ROC) and DOA, even though training is done with random patch locations. As an \textit{a posteriori} justification for choosing eyes as the location for the universal patch attack, we inspect the best location found in the square patch attack and note that the locations mostly focus around the eye region, which agrees with how humans perform face recognition \citep{keil2009look}.

\begin{table}
  \caption{Performance of the proposed approach with different test time augmentations. We compare against weak patch adversarial training and DOA \citep{Wu2020Defending} trained with a number of $C = \{4, 10\}$ mask candidates returned by the approximate search. Higher is better.}
  \label{table:results}
  \centering
    \begin{tabular}{cccccccc}
        \toprule
        &
        \multicolumn{1}{c}{Clean}
        &
        \multicolumn{2}{c}{Eyeglasses}
        &                                            
        \multicolumn{2}{c}{Square Patch}
        &                                            
        \multicolumn{2}{c}{Eye Patch}\\\hline           
        & AU-ROC & AU-ROC & AU-PR  & AU-ROC & AU-PR & AU-ROC & AU-PR    \\
        
        Ours, Basic     & $0.957$ & $0.773$ & $0.809$ & $0.742$ & $0.785$ & $0.775$ & $0.789$    \\
        Ours, Mirroring & $0.959$ & $0.812$ & $0.845$ & $0.789$ & $0.825$ & $0.791$ & $0.804$    \\
        Ours, Selection & $0.976$ & $0.831$ & $0.868$ & $0.809$ & $0.848$ & $0.884$ & $0.895$   \\
        \textbf{Ours, Both }     & $\mathbf{0.978}$ & $\mathbf{0.868}$ & $\mathbf{0.896}$ & $\mathbf{0.850}$ & $\mathbf{0.882}$ & $\mathbf{0.893}$ & $\mathbf{0.903}$    \\
        \hline
        Weak AT     & $0.886$ & $0.704$ & $0.726$ & $0.700$ & $0.722$ & $0.571$ & $0.556$    \\
        DOA, $C=4$  & $0.972$ & $0.830$ & $0.873$ & $0.836$ & $0.873$ & $0.826$ & $0.844$    \\
        DOA, $C=10$ & $0.967$ & $0.843$ & $0.878$ & $0.831$ & $0.864$ & $0.879$ & $0.886$ \\
        \bottomrule
    \end{tabular}
\end{table}

\subsection{Evasion Attack}
We run a feature adversary that starts from images $t$ with the same class identity as $x$ and maximize the objective in (\ref{eq:feature_adv}) in order to evade verification. We set the detection threshold $\delta$ such that the false positive rate is equal to $5\%$ and obtain that the attack detection rates are $25.2\%$ for our proposed approach and $19.8\%$ for DOA. While this marks evasion attacks as a weakness for both methods, the proposed approach is still able to obtain an improvement. The drop in performance can be intuitively explained by the fact that the adversary is no longer constrained by a targeted attack (i.e., impersonating a specific person).

\subsection{Indirect Anchor Attack}
Recall that our model is never exposed to pairs of real samples during training and that the two generated samples $y$ and $u$ share the same person identity and act as virtual anchors. We introduce a new type of attack in which the feature adversary \textit{intentionally} attempts to copy the features of the auto-encoded target $y$, instead of $x$. We find that this attack is almost as successful as attacking the real target $x$ and that our scheme maintains its robustness, with the AU-ROC dropping from $0.868$ to $0.867$ when we consider the most successful attack across both the regular and anchor attacks.

Even though $F$ is fully differentiable --- because of the virtual anchors $y, u$ that are used for training --- there is a reasonable suspicion of gradient masking \citep{pmlr-v80-athalye18b} on the loss surface between two real images. Since our indirect attack is almost as successful as attacking the real image of the person, we conclude that the loss surface is non-degenerate and no gradient masking occurs.

\begin{figure}
    \centering
    \includegraphics[width=1.\textwidth]{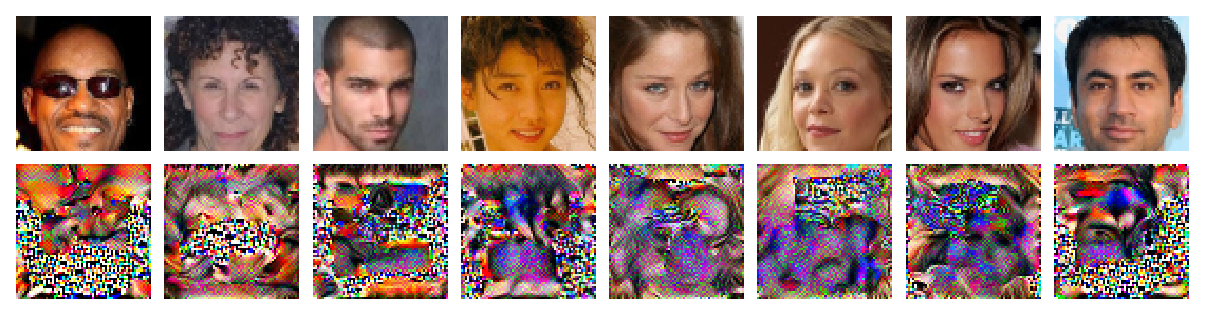}
    \caption{\textbf{Top}: Target images from which target features are extracted. \textbf{Bottom}: Images found by running unrestricted feature copy attack when starting from random noise images.}
    \label{fig:distal_results}
\end{figure}

Finally, we also perform a completely unrestricted feature copy attack (also referred to as \textit{distal} adversarial examples \citep{schott2018towards}) when starting from a pure noise image. Prior work \citep{ilyas2019adversarial} has shown that robust networks will produce samples that resemble the image from which the target features are obtained. Figure \ref{fig:distal_results} plots the results, where we notice that some distinctive features of the targets, such as eyes, mouth and forehead are present in the images. All images successfully converge to a low value of the feature distance, a further sign that gradient masking does not occur.

\section{Discussion}
Patch adversarial training using an optimal adversary at every step leads to an increase of complexity for the inner solver and raises an important practical issue: which spatial location should be chosen?  The proposed approach exhibits higher clean \textit{and} adversarial accuracy than prior work that approximately searches for the patch location. Furthermore, from the results we have obtained, it is at least apparent that randomly picking the patch location is not sufficient to ensure convergence to an accurate and robust solution by training on non-augmented data. This raises the important question of if there is a fundamental trade-off between robustness and accuracy \citep{zhang2019theoretically} when defending against patch attacks, which we consider a promising direction for future work.

Recent work \citep{Wong2020Fast, zhang2020attacks, cai2018curriculum} has investigated the adjustments that can be made to weak (single-step optimization) adversarial training in order to converge to a robust solution and has found that particular schedules of the learning rate and tuning the strength of the attacker over time improve convergence. We believe that this is a very promising topic of future research when it comes to defending against physically realizable attacks and our work contributes in this direction by enabling fast adversarial training against patches and removing one layer of complexity, i.e., searching for the optimal location. Finally, in this work we did not perform any fine tuning of the learning rates for the weight optimizer in the style of \cite{Wong2020Fast}, but we believe a larger-scale study to understand the convergence properties of patch adversarial training is an interesting research topic.

\section{Related Work}

\paragraph{Adversarial Attacks and Defenses} The closest related work to ours is \cite{Wu2020Defending}, which introduces adversarial training \citep{madry2018towards} for patch-based attacks. However, their setting is different: they apply their method to a ten-class face recognition (classification) problem, where each person identity is seen during training. In contrast, we train our model at a larger scale and test on a completely different subset of persons. \cite{Yin2020GAT:} propose an integrated training procedure for robust classification and adversarial sample detection. Our approach is similar, in that we apply adversarial training to a detection criteria, but we do it symmetrically. Specializing to patch attacks, several certified robustness methods exist for classification \citep{Chiang*2020Certified, alex2020derandomized}, but these methods still must pay a price in clean accuracy. In particular, \cite{Chiang*2020Certified} discuss the issue of providing a certificate by picking a subset of random patch locations, similar to what we do for training. Recent work \citep{Wong2020Fast, rice2020overfitting, zhang2020attacks} also investigates the dynamics of adversarial training and has shown that using a weak inner solver without extra precautions, such as manually tuning the learning rate, leads to overfitting. Our proposed approach converges without any additional tuning.

\paragraph{Face Recognition and Verification} The tasks of face recognition and verification have seen significant improvements in the recent years, starting with \cite{schroff2015facenet}. In particular, hyperspherical embeddings have seen increased popularity recently \citep{liu2017sphereface, deng2019arcface}, where they have been found to be favorable for different types of margin-based losses. \cite{zhang2016joint} tackle the problem of joint representation learning and face alignment by using the same embedding function. We decouple the two tasks and use the generative model $G$ for virtual alignment. The quadruplet loss \citep{chen2017beyond} is an extension of the triplet loss \citep{hoffer2015deep} that also uses four terms for training a similarity metrics but in a conceptually different way, since they form three pairs of samples. Regarding adversarial attacks on face recognition systems, there exists a plethora of work that shows the vulnerability of deep learning-based models \citep{sharif2016accessorize, dong2019efficient, carlini2020evading}.

\paragraph{Disentangled Latent Representations} \cite{Shu2020Weakly} recently introduced a framework for formal definitions of disentanglement, which we use to provide intuition for our approach. Learning disentangled representations has been proposed, among others, with variational autoencoders \citep{mathieu2018disentangling} and generative networks \citep{tran2017disentangled, Gabbay2020Demystifying}, in both unsupervised and supervised settings. We use the particular construction of \cite{Gabbay2020Demystifying}, but our approach works with any architecture that achieves disentanglement and can benefit from future improvements in the field.

\newpage

\medskip

\small

\bibliography{neurips_2020}

\appendix

\section{Implementation Details}
We use the same architecture as \cite{Gabbay2020Demystifying} for the generator and the encoders. $G$ is a deep neural network consisting of $3$ fully connected layers and six convolutional layers, with the final output size being an image of size $64 \times 64$ with three colour channels. We use the default values for the noise power $\sigma = 1$ (added to the content code) and regularization coefficient $\lambda = 0.001$ (regularizing the content codes) to train $G$. The generative model is trained in a non-adversarial fashion, following \cite{bojanowski2018optimizing}.

For the adversarial training of $F$, we follow PGD-based adversarial training with random patch locations. At each step, a batch of $128$ samples is selected and a number of $n_m$ samples from different classes (persons) are selected for each samples, where $n_m$ represent an integer parameter that controls the level of semi-hard example mining. We mine for hard examples before any adversarial training, but after all static augmentations (including cutout) and we pick the negative pair with the lowest feature distance out of the $n_m$ candidates.

For all samples, we apply random mirroring, random shifts of at most $5$ pixels and Gaussian cutout with probability $p=0.5$, in which a rectangle with random surface between $0.02$ and $0.2$ of the image and aspect ratio between $0.5$ and $2$ is selected, and replaced with random pixels. This is similar to the random noise initialization used in PGD-based adversarial training (and valid for us, since we assume the adversary is unconstrained in norm).

Then, for the samples where cutout is applied, we perform adversarial training with a number of $\{7, 10, 20, 40\}$ steps with step sizes $\{30/255, 16/255, 12/255, 6/255\}$, respectively. Thus, each batch consists of (on average) $128$ clean pairs and $128$ augmented pairs (where we consider adversarial training a form of augmentation). 

For weak adversarial training, we attack all samples in the batch and select a random patch location with random surface between $0.02$ and $0.1$ of the image and aspect ratio between $0.5$ and $2$. We use a smaller upper bound for the surface of the patch, since we find that training with $0.2$ does not decrease the test loss significantly after $2000$ steps but still exhibits the same overfitting phenomenon, thus a good solution cannot be obtained even with early stopping.

For DOA, we attack all samples in the batch and search for a square patch of size $20 \times 20$ pixels (which matches the average patch size and aspect ratio for the proposed approach), with a stride of $5$ pixels, since searching across a range of aspect ratios is computationally intractable. For both our method and DOA, we find that the training is not critically sensitive on the number of steps, thus we train with $10$ steps and step size $16/255$ for both. We train all models for $20000$ steps (or, for weak adversarial training, until divergence occurs) using an Adam optimizer with learning rate $0.001$ and default values $\beta_1 = 0.9$, $\beta_2 = 0.999$.

\section{Additional Results}

\subsection{Performance without Adversarial Training}
Table \ref{table:clean_performance} shows the results for our approach without any adversarial training (but with both test time augmentations included), where the impact of weak adversarial training can be measured and it is revealed as a necessary component for a robust solution (but not for clean performance).

\begin{table}[h!]
  \caption{Performance of our proposed approach with and without added adversarial training. Note that only with $n_m = 2$, a very weak form of semi-hard example mining, our method converges to a stable solution (in terms of clean accuracy) without adversarial training.}
  \label{table:clean_performance}
  \centering
    \begin{tabular}{cccccccc}
        \toprule
        &
        \multicolumn{1}{c}{Clean}
        &
        \multicolumn{2}{c}{Eyeglasses}
        &                                            
        \multicolumn{2}{c}{Square Patch}
        &                                            
        \multicolumn{2}{c}{Eye Patch}\\\hline           
        & AU-ROC & AU-ROC & AU-PR  & AU-ROC & AU-PR & AU-ROC & AU-PR    \\
        
        Without AT     & $0.977$ & $0.590$ & $0.669$ & $0.695$ & $0.757$ & $0.657$ & $0.608$ \\

        With AT     & $0.978$ & $0.868$ & $0.896$ & $0.850$ & $0.882$ & $0.893$ & $0.903$ \\
        \bottomrule
    \end{tabular}
\end{table}

\subsection{Performance under Random Noise}
We evaluate the proposed approach against two types of random perturbations. In the first we consider a random noise attack, where an intruder adversary does not run a gradient descent algorithm to find the value of the perturbation, but instead generates a number of $1000$ random noise patterns, fills the perturbation mask with them and then selects the one that minimizes feature space distance to the target. The detection AU-ROC under for a random eyeglasses attack suffers a minor drop from $0.978$ (clean) to $0.973$ when accumulated over $100$ targets and $900$ intruders for each target, thus we consider this another sanity check against gradient masking.

The second perturbation is additive random noise applied to the entire image. This investigates the robustness of the model under non-adversarial global perturbations. We add uniform pixel noise with values between $[-10/255, 10/255]$ and obtain an AU-ROC of $0.965$ compared to a clean value of $0.978$, indicating that the proposed scheme is robust to non-adversarial uniform (across the entire image) noise as well.

\subsection{Weak Adversarial Training}
\begin{figure}[h!]
    \centering
    \includegraphics[width=0.9\textwidth]{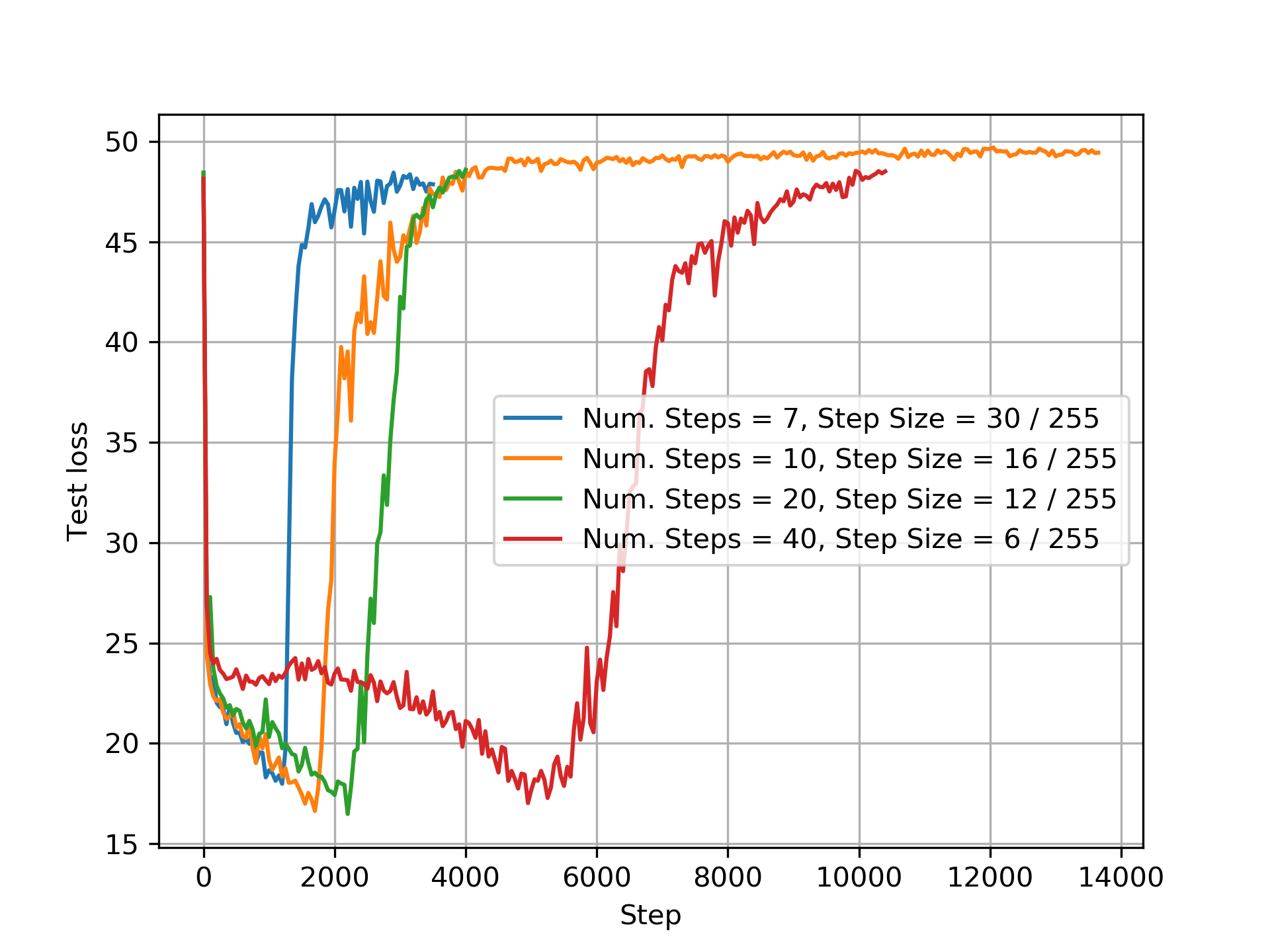}
    \caption{Test loss of weak adversarial training, where random patch locations are selected at each step. Critical overfitting occurs for all tried step sizes, indicating that random patch location is not sufficient to ensure convergence to a robust solution.}
    \label{fig:val_loss_weak}
\end{figure}

Figure \ref{fig:val_loss_weak} shows the critical overfitting phenomenon highlighted in the main text happening for all tested combinations of adversarial steps and step sizes. We use early stopping on a validation set of $100$ persons to pick the best performing model and use it as a reference (the \textit{Weak AT} line in our main results).

\section{Details on Adversarial Attacks}

To implement our attacks, we adopt the hyperbolic tangent formulation of \cite{carlini2017towards}. That is, the perturbation $\delta$ is the solution to the optimization problem

$$
\min_\delta \norm{F(\tanh{(x + \delta \cdot \mathcal{M})}) - F(x_t)}_2^2
$$

\noindent where $x$ is the input image converted with the hyperbolic arc-tangent function, and we also use appropriate scaling factors to ensure the resulting image belongs to $[0, 1]$. For evasion attacks, we use the same objective, but maximize it instead (minimize its negative). For the eyeglasses and square patch attacks, we use a number of $1000$ iterations with learning rate $0.01$. To pick the optimal patch location, we first run a number of $20$ iterations at $0.1$ learning rate for all possible patch locations (with a stride of $5$ pixels), pick the location and then restart optimization. For the universal eye patch, we use $5000$ iterations at $0.005$ learning rate. For the distal attack, we use $2000$ iterations at learning rate $0.01$.

\begin{figure}[h!]
    \centering
    \includegraphics[width=1.\textwidth]{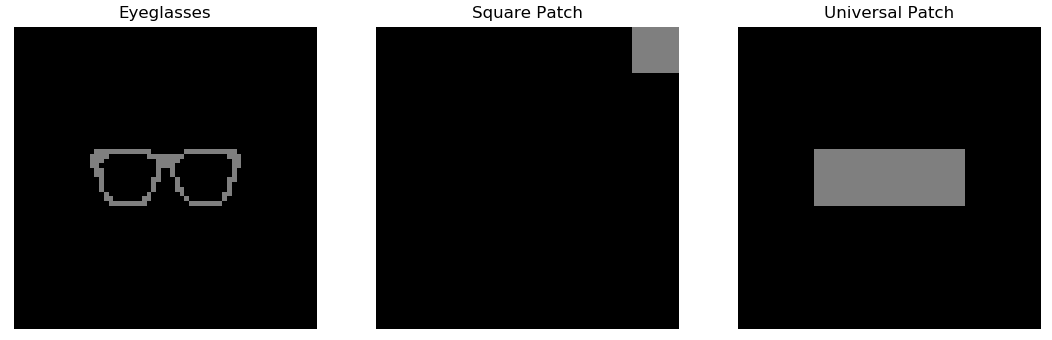}
    \caption{The three perturbation masks an adversary is allowed to operate on. \textbf{Left}: Eyeglasses. \textbf{Middle}: Square patch (location varies). \textbf{Right}: Universal eye patch.}
    \label{fig:mask_types}
\end{figure}

Figure \ref{fig:mask_types} shows examples for the three perturbation masks used in our evaluation. For the eyeglasses perturbation mask, we use a public domain image available at \href{https://pngio.com/images/png-a1902002.html}{https://pngio.com/images/png-a1902002.html}, which is reshaped to an image of size $32 \times 16$ and placed in the same location for all images.

\begin{figure}[h!]
    \centering
    \includegraphics[width=0.7\textwidth]{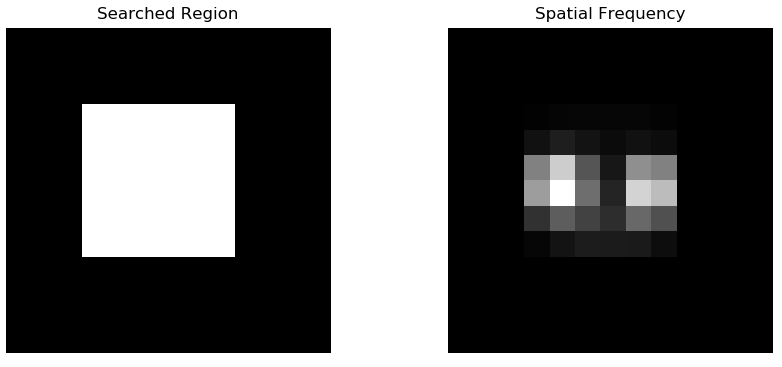}
    \caption{\textbf{Left}: Region in which the adversary searches for the best patch location. \textbf{Right}: Spatial distribution of optimal patch locations found. Lighter color indicates higher frequency.}
    \label{fig:best_mask_location}
\end{figure}

For the square patch attack, we pick a $10 \times 10$ pixel mask in the region of the image, searching with a stride of $5$ pixels in both dimensions. Figure \ref{fig:best_mask_location} plots the search region, as well as the distribution of near-optimal locations for the attacks, where a preference for the eyes can be noticed. For the universal eye patch attack, we place a rectangular patch of size $32 \times 12$ in the eye region. To find the eye region, we average the eye center annotations in the CelebA dataset.

\begin{figure}[h!]
    \centering
    \includegraphics[width=1.\textwidth]{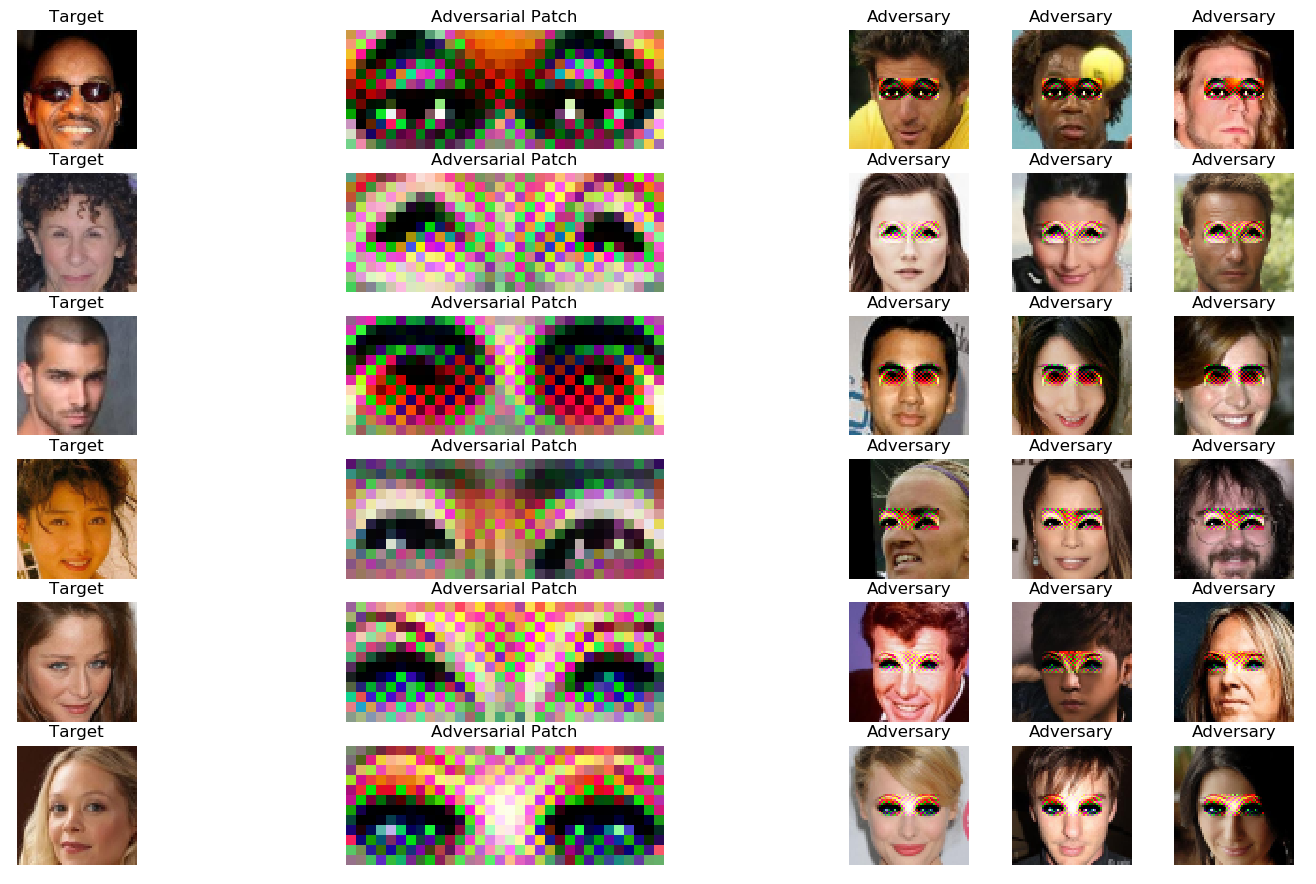}
    \caption{\textbf{Left}: Target images used as anchors in feature space. \textbf{Middle}: Adversarial eye patches (up-sampled) after training on $100$ intruders. \textbf{Right}: Adversarial eye patches applied to other intruders (not in the training set, randomly picked).}
    \label{fig:examples_eyepatch}
\end{figure}

Figure \ref{fig:examples_eyepatch} shows several examples of trained universal eye patches, the target images, and a set of test adversarial samples (not used for training the patch). It can be noticed that the eye patch is semantically meaningful and captures features of the target person.

\newpage

\section{More Examples of $x, y, t, u$ Quadruplets}
\begin{figure}[h!]
    \centering
    \includegraphics[width=1.\textwidth]{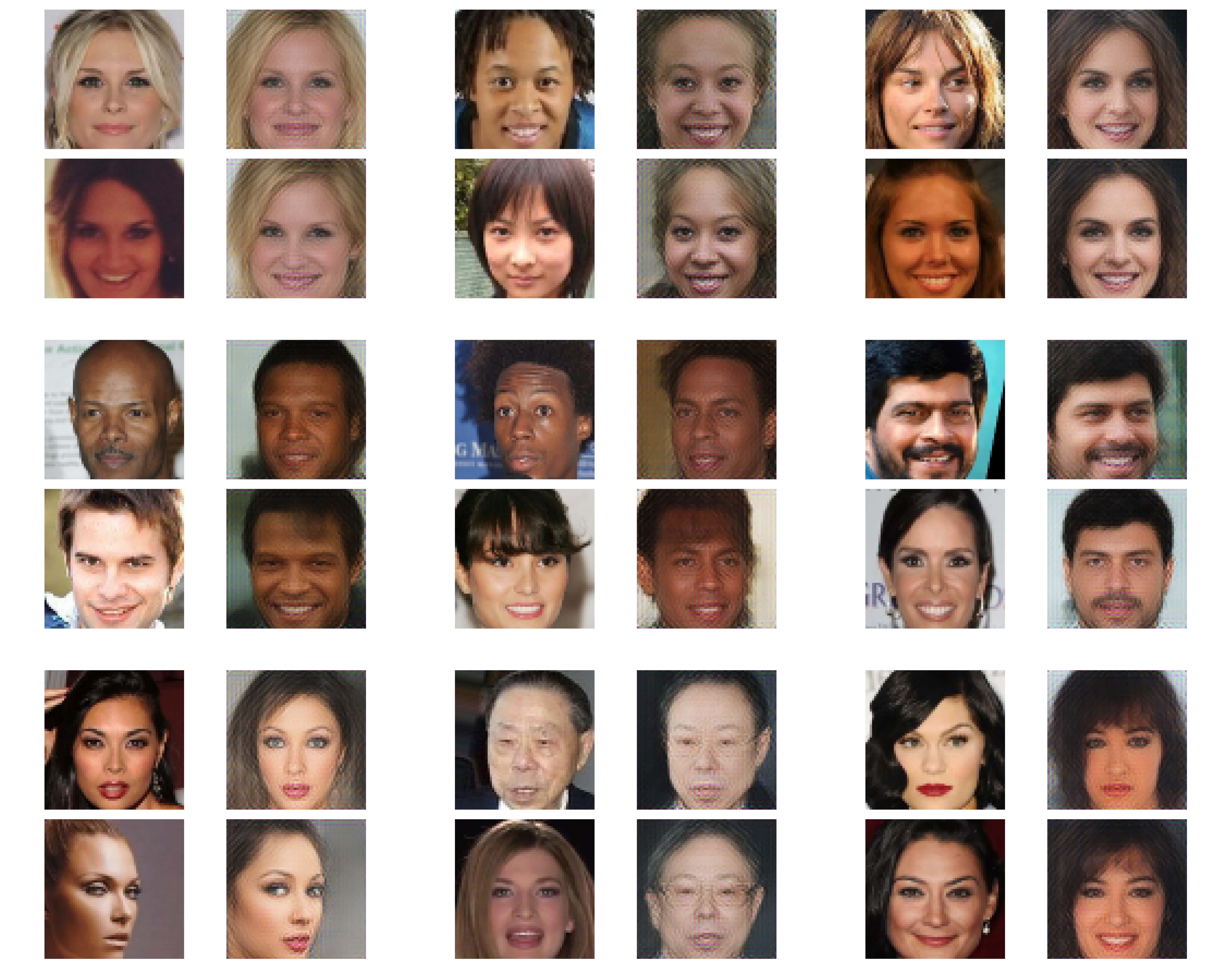}
    \caption{More examples of quadruplets $x, y, t, u$. In each sub-block of $2 \times 2$ images: upper left -- $x$, upper right -- $y$, lower left -- $t$, lower right -- $u$. $(x, y$) is the positive pair and $(t, u)$ is the negative pair.}
    \label{fig:more_quadruplets}
\end{figure}

Figure \ref{fig:more_quadruplets} shows sets of random (not cherry picked) quadruplets. In some cases, it can be noticed that the class identity of $y$ (and, implicitly, $u$) is slightly distorted under human inspection, highlighting its use as a virtual anchor for learning similarity.

\end{document}